\documentclass[twocolumn]{article}
\usepackage[a4paper,margin=1in]{geometry}
\usepackage[hyphens]{url}
\usepackage{graphicx}
\usepackage{caption}
\usepackage[utf8]{inputenc}  
\usepackage[T1]{fontenc}    
\usepackage{lmodern}        

\usepackage{amsmath, amssymb, mathtools, amsthm}

\usepackage{algorithm}
\usepackage{algorithmicx}
\usepackage{algpseudocode}

\usepackage{multirow}
\usepackage{subcaption}

\usepackage{pgfplots}

\newtheorem{definition}{Definition}

\title{Towards Intention Recognition for Robotic Assistants Through Online POMDP Planning}

\author{
    Juan Carlos Saborío\textsuperscript{\rm 1} \and
	Joachim Hertzberg\textsuperscript{\rm 1,\rm 2}
}
\date{
	\textsuperscript{\rm 1} German Research Center for Artificial Intelligence (DFKI)\\Hamburger Straße 24, 49084 Osnabrück, Germany\\
    \textsuperscript{\rm 2} Institute of Computer Science, University of Osnabrück\\Wachsbleiche 27, 49090 Osnabrück, Germany\\
    Email: \texttt{juan.saborio@dfki.de, joachim.hertzberg@uos.de}
}

\begin{document}
\maketitle

\begin{abstract}
Intention recognition, or the ability to anticipate the actions of another agent, plays a vital role in the design and development of automated assistants that can support humans in their daily tasks. In particular, industrial settings pose interesting challenges that include potential distractions for a decision-maker as well as noisy or incomplete observations. In such a setting, a robotic assistant tasked with helping and supporting a human worker must interleave information gathering actions with proactive tasks of its own, an approach that has been referred to as \emph{active goal recognition}. In this paper we describe a partially observable model for online intention recognition, show some preliminary experimental results and discuss some of the challenges present in this family of problems.
\end{abstract}

\section{Introduction}
Imagine a busy factory setting where workers and robots collaborate towards shared goals. For example, if a worker is required to inspect a machine, they will initiate a series of steps - a plan - to systematically determine which part or which component is faulty and whether it needs to be replaced. A robotic assistant might support this worker and attempt to provide tools and parts as needed, but the robot and the worker need a shared understanding of the problem and the robot must be able to assess the worker's needs.  Another worker may be tasked with assembling a series of objects by picking parts from nearby containers, and another robotic assistant may be able to help by observing the worker's activities, estimating whether some containers need to be restocked and bringing tools the worker may need soon.  In both cases the worker and the robot have shared goals and a shared understanding of the environment and the problem, but the worker may focus exclusively on their chores while the robot performs support tasks to avoid or minimize issues.  Such robots must be able to receive and assess information about the worker's activities and the environment in order to act appropriately, and consider that this information may not be accurate or even complete.

These scenarios raise the following question: how can an automated assistant support a human worker in a collaborative environment, without explicit instructions, as well as plan and act with noisy and incomplete observations? At least some of these components have been addressed in the past, and we have learned lessons from automated assistance in e.g., nursing homes \cite{PineauPearl2003, Hoey2010}.  Further improving recognition, autonomy and resilience in complex domains leads to a series of important, interrelated challenges collectively addressed by the ``Plan, Activity and Intent Recognition'' (PAIR) community.

Although the different types of recognition in PAIR share principles and methods, such as some type of knowledge base (e.g. a plan library) and some way to generate hypotheses over prospective answers, distinctions exist in that \emph{activities} commonly refer to singular actions, \emph{plans} (or goals) are the best explanation for a sequence of observed actions and an \emph{intent} may be seen as a \emph{motivation}, or rather subsets of goal conditions or possible future actions \cite{FreedmanZilberstein2019}. From this point on we'll refer to the agent performing the PAIR tasks as the \emph{observer} and to the agent being observed as the \emph{target} (the robot and the worker respectively in our motivating examples).

Plan recognition can also be seen as a \emph{reverse} planning problem, where the input is a sequence of actions and the output is a goal that explains this sequence \cite{RamirezGeffner2010, RamirezGeffner2011, Sohrabi2016}. Intent and goal recognition may also be addressed using similar approaches, but instead focusing on the goal conditions being pursued by the target.  Many such techniques, however, only observe the target's behavior in order to approximate the likelihood of a plan or goal hypothesis. Alternatively, Active Goal Recognition (AGR) entails not just identifying goals but also interacting with the environment or the target \cite{Amato2019}, balancing task planning with information gathering actions that yield observations about the environment and the target's behavior. In such scenarios an assistant (technically no longer \emph{just} an observer) collects information and acts in ways that are informed by the target's behavior, since they have shared goals that require interaction or cooperation (as in our examples).

In this paper we propose a POMDP approach to model active intention or goal recognition tasks, and as suggested in recent papers, we solve them using an online, anytime POMDP planning framework that does not require an explicitly factored model \cite{SaborioHertzbergLNAI19, SaborioHertzbergUAI19}. In addition, this POMDP planner adopts an approach called ``relevance-based planning'' which yields promising results in such problems.  We argue that the interplay between intention recognition and relevance estimation (or an equivalent) not only produces promising results but it highlights the impact of improved autonomy in action selection, particularly when solving problems that require interleaved planning and acting, and opens the door for future improvements.

\section{Related Work}
A substantial amount of prior work in plan recognition adopts a passive perspective in which the behavior of a target agent is observed, but the observer does not directly interact with the environment \cite{RamirezGeffner2010, RamirezGeffner2011, Geib2015, Sukthankar2014}. Plan libraries are a popular approach that contain either precomputed plans, or a grammar that can construct possible plans \cite{GeibGoldman2009, Geib2015, Rafferty2017}. The challenge then is estimating the probability of some given plan in the library being the current one, given observations about the target's behavior or their activities. Alternatively, the planning approach to plan recognition attempts to identify plans, goals, or intentions by observing the target's actions and generating, as well as evaluating, the space of possible target plans \cite{RamirezGeffner2010,Sohrabi2016,FreedmanZilberstein2017}. While there is a strong reliance on classical planners that assume determinism and full observability, some work has also been done to address missing information \cite{Sohrabi2016} and to extend goal recognition to probabilistic and partially observable domains \cite{RamirezGeffner2011}.

As our motivating examples suggest, we are interested in problems that fall within AGR.  Building on previous work in partially observable domains, a detailed POMDP representation for states, actions, transitions, rewards, and observations has been proposed that combines the observer and target variables in factored structures \cite{Amato2019}. Our approach also uses shared POMDP states with target and observer variables but no explicit distiction is made, and instead we adopt a flat representation (without factoring) and model all action effects as part of the POMDP dynamics (or state transitions).  We argue this achieves a similar level of expressiveness by allowing the target to behave independently of the observer, and restricting the observer's ability to obtain perfect information, while also maintaining a combination of fully and partially observed variables. With additional effort perhaps our model could be converted to the factored AGR POMDP, but we think a flat representation is more straightforward.  The authors used the point-based planner SARSOP to solve the AGR problems, but point out the benefits of using online POMDP planning. Point-based planners expand and evaluate large sections of the belief space and are restricted to fairly small domains, even if they are neatly factored. Additional work that directly addresses AGR includes a mixed-observability approach \cite{MassardiBeaudry2021}, but again the authors recommend considering an online approach to deal with limitations including large state spaces and combinatorial explosion.

POMDP planning on its own has a vast amount of literature that can perhaps be quickly summarized as a series of substantial improvements in approximation, using different techniques ranging from direct value iteration to grid-based and compression methods \cite{SmallwoodSondik73, CassandraAAAI94, CassandraUAI97,Bonet2002,PoupartBoutilier2002,TheocharousMahadevan2010}.  Many commonly used planners follow point-based methods \cite{Smith2004,PineauPBVI2003,PineauPBVI06,SpaanVlassis2005,SARSOP}, but these tend to be limited to small problems, often in factored form.

Online POMDP planning algorithms scale much better by focusing on the current belief state and computing value-updates often using sampling or generative approaches \cite{SilverPOMCP, Somani2013_DESPOT, GPS-ABT, POMCPOW}.  POMCP is a well-known POMDP planner based on Monte-Carlo Tree Search (MCTS) \cite{SilverPOMCP}, that continues to obtain good performance even in larger problems, with the limitation that it may produce low-performing worst-case results. These algorithms may also exploit prior knowledge to improve performance, but this is often in the form of extensive domain knowledge, or highly detailed a-priori action preferences. An approach called relevance-based planning builds on the structure of POMCP, but improves performance by introducing online methods that 1) generate action selection preferences autonomously, and 2) reduce dimensionality through observation-based criteria \cite{SaborioHertzbergLNAI19, SaborioHertzbergUAI19}. We use this planner to obtain preliminary experimental results and compare its performance with standard POMCP, which also helps illustrate the connection between improved agent autonomy and performance in intention recognition tasks.

\section{Brief Background on POMDPs}
Let $S$ and $A$ be finite sets of states and actions, $T(s,a,s') = p(s'|s,a)$ the transition probability to state $s'$ (with $s \in S, a \in A$) and $R$ a set of real-valued rewards.  The tuple $\langle S,A,T,R \rangle$ defines a fully observable MDP and $0 \leqslant \gamma \leqslant 1$ its discount factor.  In partially observable domains the agent maintains an internal belief state $b \in B$, where $b(s)$ is the probability of $s$ being the current state.  This probability is updated from observations $\omega \in \Omega$, received with probability $O(s',a,\omega) = p(\omega | s',a)$.  A POMDP is defined as the tuple $\langle S,A,T,R,\Omega,O \rangle$, and the sequence $h_t = (a_0, \omega_1, \ldots, a_{t-1}, \omega_t)$ is the \emph{history} at time $t$.  A full POMDP policy maps beliefs to optimal actions, but when planning and acting online only the current belief and the next best action are considered.

An action for a belief is found by approximating the optimal state-value function:

\begin{equation}
\begin{aligned}
v_*(b) = \max_a & \left[ \sum_s b(s)R(s,a,s') ~ + \right. \\
		 & \left. \gamma \sum_\omega O(s',a,\omega) v_*(b') \right]
\end{aligned}
\end{equation}

which reflects the expected return (sum of discounted rewards) when starting in $b$ and pursuing the optimal policy. The optimal action-value function $q_*(b,a)$ represents the value of executing $a$ in $b$ and then pursuing the optimal policy. In our approach both the observer and the target are represented in the POMDP states (and variations thereof in the belief state), so states where the target achieves goal conditions are the result of transitions where the observer acted correcly in advance. Such transitions yield positive rewards which, succintly put, transform intent recognition into an action-value maximization problem.

\section{Model Description}
Similar to previous work in AGR, c.f. \cite{Amato2019}, we propose a POMDP model where both the agent and the observer share a state representation (i.e. through state variables) but the target consists of a small stochastic model that simulates target actions independently from the observer, and is capable of generating reward signals during state transitions. This independence as well as the generative aspect of target simulation make it easy to represent active intention recognition tasks as unfactored POMDPs that can be solved online, in a fairly straightforward manner.

More formally, let $S_O$ be the set of state variables that represent the observer and $S_T$ the set of state variables that describe the target.

The target is described by a policy $\pi_T$, which is a solution to a stochastic process $\langle S,\Omega_T, T, R_T \rangle$ defined by:

\begin{itemize}
\item $S_T$ is a finite set of target states
\item $\Omega_T$ is a finite set of target actions
\item $T(s_T,\omega_T,s'_T)$ is the transition probability from $s_T$ to $s'_T$ after executing action $\omega_T \in \Omega_T$
\item $R_T(s_T, \omega_T, s'_T)$ is a set of real-valued rewards
\end{itemize}

In other words, $\pi_T$ is a target simulator that can take a target state as an input and generate the state resulting from its transition model. We then incorporate this into a POMDP, redefined as:

\begin{itemize}
\item $S = S_O \times S_T$ the state space described by all possible observer and target variables. States are the result of $S_O \cup S_T$ but not explicitly factored.
\item $A$ the set of observer actions.
\item $\Omega = \Omega_O \cup \Omega_T $ is the set of observations that combines non-target observations $\Omega_O$ and target activities $\Omega_T$. Again, not explicitly factored.
\item The transition model $T = T(s,a,s')\pi_T(s_T)$, such that:
	\begin{itemize}
	\item Target variables $s'_T = s' \cap S_T$ are determined by $\pi_T(s_T)$
	\end{itemize}
\item $R = R(s,a,s') + R(s_T,a,s'_T)$ is a real-valued reward function that includes rewards from the observer's actions and rewards from the target's actions.
\end{itemize}

Structural changes include the addition of $\pi_T$, a priori non-deterministic policy that simulates the target's behavior and constitutes a generative replacement for a plan library, as well as the target's rewards when transitioning to states that meet goal conditions.  This approach to target behavior lends itself well to online, generative POMDP planning approaches such as those based on MCTS. The target's behavior is perceived through observations $\omega_t \in \Omega_T \subseteq \Omega$, which are target activities. For the observer (the planning agent), no explicit distinction is made between actions that modify the domain and those that gather information; during and after planning the agent simply follows a converging action-selection function (such as UCB1 which minimizes regret).

This creates a more or less standard, unfactored POMDP with the added complexity of a ``target'' element, represented by additional state attributes that are beyond the planning agent's control and that change over time. The target acts independently but shares goal conditions with the observer, and the observer receives noisy information through its sensors. Since the target may not be able to achieve their own goal conditions without the observer's participation (e.g., the worker may not have the right parts or tools available), the observer needs to act accordingly in order to reach those particular states.

We can define the target goal conditions as $G_T$, which are given in advance as part of the stochastic model $\pi_T$ (for instance assemble objects, replace damaged components, etc.) and the observer's own goal conditions as $G_O$ although these arise implicitly from the POMDP dynamics as is usually the case. In other words, the observer is always concerned with maximizing long-term, discounted return; high returns are obtained as a result of the observer's own actions which may lead to transitions in $\pi_T$ that satisfy conditions in $G_T$. We then define our intention recognition approach as:

\begin{definition}[Online Active Intention Recognition]
\label{def1}
An intention recognition problem is represented by the tuple $\langle \Sigma, s_0, \pi_T, G_T \rangle$ which consists of a planning domain $\Sigma$ (our POMDP), an initial state $s_0$, a stochastic target model $\pi_T$ and a set of target goals $G_T$.
\end{definition}

In our examples, the worker cannot succeed without the assistant's effective intervention. In other words, the observer must interact with target in directed ways that satisfy active goal conditions in a timely manner. This means that the optimal policy in such a problem is one where the observer performs the necessary actions to maximize return from its own actions, some of which include interacting with the target in a directed, informed way in order to guarantee their shared goals are satisfied and terminal states reached.

\subsection{Relevance Estimation and Intention Recognition}
Relevance-based planning is an approach to improve the performance of planning under uncertainty where agents (such as robots) must complete tasks in complex domains that 1) Have incomplete and noisy information, 2) Provide many interaction opportunities, 3) Require quickly identifying suitable goals and subgoals to pursue \cite{SaborioHertzbergLNAI19, SaborioHertzbergUAI19}. The methods are presented as improvements to POMDP planning, particularly with respect to generative algorithms such as MCTS.

Functionally, it consists of ``Partial Goal Satisfaction'' (PGS) and ``Incremental Refinement'' (IRE). PGS is a method that estimates goal proximity by awarding positive or negative points when goal conditions are met or broken, respectively, upon state transitions, leading to an improved rollout policy and a reward bonus during simulation.  IRE is a dimensionality reduction method that estimates a \emph{relevance} value, as a function of ``features'' (the elements of the domain that provide actions to the agent).  The contribution or reward from the actions in each feature are aggregated, weighed and scaled resulting in a value that represents their impact in problem solving, enabling the agent to focus on relevant actions and avoid those with poor or non-goal-related outcomes. An example in the AGR setting could be a robot preferring to gather information about tools or parts that a worker might need soon, and avoiding those that appear to play no role in the perceived plan.

To the best of our knowledge, existing POMDP-planning based approaches for active intention recognition (or similar) use out-of-the-box planners that do not exploit the structure of such problems, or use point-based planners limited to very small problems. The appeal of the relevance-based approach is that it may already improve performance with little more than the information already available when specifying an active intention recognition problem (as per def. \ref{def1}).

We incorporated the PGS component into the intention recognition framework by assigning points to the goal conditions in $G_T$, and negative points for conditions that violate $G_T$. The total score is computed by the sum of these points in some given state or POMDP history, which results in the reward shaping function:

\begin{equation}
F(h_t, h_{t+1}) = \gamma \phi(h_{t+1}) - \phi(h_t)
\end{equation}

where

\begin{equation}
\phi(h) = \alpha \mathfrak{p}(h)
\end{equation}

uses the PGS point scoring function $\mathfrak{p}(h)$ for history $h$ and $\alpha$ is a real-valued scaling factor. During simulation, an agent may perceive a significant reward bonus on transitions that meet goal conditions, which then backpropagates to make the originating action stand out even in relatively long planning horizons. Additionally, we use the PGS rollout policy:

\begin{equation}
A(h) = \arg \max_a \mathfrak{p}(h \cup \{a,\omega\})
\end{equation}

which generates a subset of actions, during MCTS rollouts, that lead to histories with maximal PGS points; these are histories that meet (the most) goal conditions. Notice that these points are cummulative, so an optimal (goal) state is one with the maximum possible points.  We also incorporated but did not thoroughly test the IRE component. With additional work, however, we anticipate it may help reduce the size of AGR problems where an agent has large sets of actions or tasks and many do not contribute to problem solving (e.g., many parts to bring but the worker does not need them).  These methods were designed to improve performance in large, unfactored POMDPs and as such they make some structural assumptions, but it could be argued such assumptions are valid across many robotics tasks including AGR.

\subsection{Challenges in Active Intention Recognition}
As previously stated, the passive approach to intention recognition is primarily based on observing and classifying the target's behavior, and may account for missing observations and perception uncertainty. When the observer is allowed to actively participate in the problem, however, it can receive a much broader variety of observations coming from its extended action set, many of which may not yield information about the target. Obtaining sufficient information becomes part of the deliberation process and we argue that a successful model should account for this by providing a suitable array of information gathering actions. In other words, active intention recognition may in fact require active or directed observation.

Particularly when using POMDPs, informative observations are crucial to correctly transform the belief state and approximate the true, ``real-world'' state.  This task however may become challenging if an \emph{active} observer is engaged in tasks that limit or prohibit observing key predictive behaviors, and outright insurmountable if such information becomes inaccessible. For example, in our object assembly example, two different objects that require different tools may consist of almost entirely the same parts, except for one which the robot may not observe when performing independent work. In this case, the ability to directly observe properties of the domain, such as the object being assembled, may be necessary.

An additional challenge is how the observer, or planning agent, perceives rewards for useful actions over a potentially long horizon and the immediate reward for most actions might be 0, or even negative. This appears to be a characteristic of active intention recognition because, by definition, the observer acts in anticipation of the target's actions and only later are reward-generating conditions satisfied. For example, in the assembly task the robot's goal is not simply to restock all parts, but to restock parts that \emph{will be needed} by the worker to assemble objects. The robot can bring more parts at any time (at a cost) even if they are not needed, or perhaps all necessary parts are already present and no further action is required. The worker, on the other hand, generates positive rewards when an object is assembled successfully (a condition in $G_T$) which may be received by the observer several steps after the relevant action was performed (e.g., restocking the correct part). This situation makes planning over large spaces particularly difficult, as any given observer action may have many different outcomes across over some finite horizon that make it difficult estimate its true action value. This problem is inherent to POMDP planning but possibly amplified with such problem dynamics.

Recognizing behavior and intentions from both direct and indirect observations as well as acting with delayed rewards constitute significant hurdles for active intention recognition.  In summary, our proposal is to utilize a model that is suitable for online, generative planning and that integrates well with existing contributions that improve the observer's ability to generate context-driven preferences, such as the relevance-based methods.

\section{Preliminary Results}
We modeled the two domains introduced in the beginning of the paper using the online, active intention recognition approach.  The resulting problems were solved with two MCTS-style POMDP planners: RAGE (which implements the relevance-estimation approach) and POMCP, a uniformly random planner commonly used as a state-of-the-art baseline. Although much more work is pending, current results show promise in solving intention recognition tasks and their resulting, large, unfactored POMDPs in an online manner.

In both problems we adopted a triple of the form $\langle a, o, r \rangle$ for target observations, where $a \in \mathcal{A}$ is a target action, $o \in \mathcal{O}$ is an object and $r \in \mathcal{R}$ is the result of the target performing $a$ on $o$. The set of target activities in $\pi_T$ is therefore $\Omega_T = \mathcal{A} \times \mathcal{O} \times \mathcal{R}$, and the state space the cartesian product of all possible activities and state variables of all objects. Internally such structure is irrelevant for the POMDP planner, but it makes the problems easier to model, understand and discuss and may in fact transfer to other intention recognition domains.

\subsection{Maintenance scenario}
In the maintenance problem, a worker must inspect a machine consisting of a compartment and a circuit board. The worker may visually inspect the compartment to determine if it is loose, and if it is proceed to tighten it with a screwdriver. The circuit board has a relay, which may or may not be damaged. To inspect it the worker needs a multimeter that shows the relay status, and if damaged the worker proceeds to replace it. The worker actions are $\mathcal{A} = \{$none, inspect, replace, screw$\}$, the objects $\mathcal{O} = \{$board, compartment$\}$ and the possible results $\mathcal{R} = \{$OK, NOT OK, NONE, FAIL$\}$. If result is OK the component works as intended, NOT OK means that it is faulty, FAIL that the action could not be executed (e.g. tool was missing) and NONE that the action has no associated outcome. We also added 3 additional tools that do not help the worker but the robot may nevertheless bring. All tools have a cost, which represent the amount of steps or turns the robot needs to get there, collect them and bring them back to the worker. During this time the worker continues to act according to $\pi_T$ so the robot may miss several observations. The problem is solved when all components have been inspected and determined to be OK.

Transitions in $\pi_T$ are stochastic but in general move the worker from $a =$ none to $a =$ inspect and with $p = 0.5$, to either $o =$ board or $o =$ compartment. Depending on $r$ the worker moves back to $a =$ none and inspects again or proceeds to replace or screw respectively. Visual compartment inspections succeed with probability, as do using the screwdriver but the multimeter always returns the correct relay status. The success probabilities are directly proportional to the worker's expertise level, another input variable. Also, the worker does not produce any rewards in this problem.

The robot actions are:
\begin{itemize}
\item ``Perceive'', which returns an $\langle a,o,r \rangle$ observation
\item ``Bring'' the screwdriver, multimeter, relay or any of the 3 unnecessary tools
\end{itemize}

A tool is "needed" when the worker transitions to a state that requires it (e.g. screw compartment, measure relay, replace relay).

The reward distribution is:
\begin{itemize}
\item -10 if a tool is brought and not needed
\item -2 if a tool/part is needed and missing
\item -0.5 perceive
\item +5 if a tool is brought and is needed
\item +10 terminal state
\end{itemize}

And the PGS point distribution is $+1$ for each tool that is needed and present, and $-1$ for each tool that is either needed but not present, or present but not needed.

In this problem, the worker depends on the robot's intervention to provide tools and a spare relay if necessary. At the very least it will require a multimeter to measure the current relay, and if both components are already in good condition no other actions are required. In the beginning however the robot does not know the true state of the components or the worker $\langle a,o,r \rangle$, both of which are obtained through observation.  The reward and PGS distribution also causes the robot to attempt to provide tools and parts at the time when they are needed, to prevent bringing them too soon or too late and to avoid bringing unnecessary tools.

\subsection{Assembly scenario}
The maintenance scenario requires the worker to assemble two toy trucks (one ``red'' and one ``blue''), each of which is composed of parts such as a cabin, chassis and wheels. A part is taken from a set of storage containers within reach of the worker; each container has only one type of part but there may be several containers. Once all parts are in place, a truck needs one of two types of glue to be fully assembled, and afterwards the worker moves on to assemble the next truck. The problem is finished when all trucks are successfully assembled. In the layout we used, the worker actions are $\mathcal{A} = \{$none, assemble, wait$\}$, objects are $\mathcal{O} = \{$chassis, wheels, blue cabin, yellow cabin, red cabin, container$\}$ and the results $\mathcal{R} = \{$OK, FAIL$\}$. Same as before, FAIL means a worker action could not be completed (missing parts, missing glue). If the worker needs a part from an empty container, the container becomes ``needed''.

The worker simulation $\pi_T$ cycles the worker in a loop: in $a =$ none, try to assemble the next part of the current truck. If OK, mark as assembled and repeat, otherwise wait and go to $a =$ none. If all parts are assembled, attempt to glue and if OK, move to the next truck and repeat. This means that the robot must continually monitor both the environment and the worker activities to minimize the amount of failures and avoid forcing the worker to wait. The worker on the other hand may be unable to complete their task without the robot's help. In this setting $\pi_T$ generates rewards of $-5$ when a part or the correct glue is missing, $-2$ every time they wait and $5$ when a truck is successfully assembled.

The robot has the following actions and observations:
\begin{itemize}
\item ``Perceive worker'', which generates an $\langle a,o,r \rangle$ observation
\item ``Inspect'' the current object/truck as well as each individual container (a total of 6). With a probability proportional to its sensor accuracy, the robot perceives the correct status (empty, not empty) or truck type and otherwise, it receives a noisy reading.
\item ``Restock'' any container (out of 6)
\end{itemize}

The reward distribution is:
\begin{itemize}
\item $-0.5$ perceive and inspect
\item $-2$ restock any part, bring any glue
\item $-5$ if restocking exceeds storage capacity
\end{itemize}

PGS points are: $+1$ for each completed truck, $-0.5$ for each incomplete truck and $-1$ for each container that is empty and needed. By attempting to maximize points, the robot will focus on restocking necessary containers by observing them directly and attempt to provide the correct glue for each truck quickly, which involves receiving observations about both the worker and the current truck.

This is a fairly intricate problems with complex dynamics at play, such as the interplay between the worker and the robot (and the aforementioned challenges such as delayed rewards). When using an MCTS planner like POMCP where belief state are approximated with particles, it is also important to prevent particle deprivation by generating state samples. We accomplished this in a non-trivial way by introducing 3 types of random variations: modifying a random container, altering the perceived worker $\langle a,o,r \rangle$ and changing the current truck type. Each transformation is validated by comparing the resulting sample with a real-world observation. For example, if the truck type was observed to be $1$ or container 3 was observed to be empty, no contradictory transformations can be accepted.

\subsection{Performance in Active Intention Recognition}
The following plots show the discounted returns collected by the observer in the two active intention recognition problems, as a function of an increasing amount of MCTS simulations per step for a maximum of 100 steps, after which the episode ends. The results were averaged over 100 episodes. It is worth mentioning that the reward bonuses in RAGE apply only during the planning and simulation stages; once an action is selected and executed (in this case, also simulated) all agents perceive rewards in the same, problem-defined range.

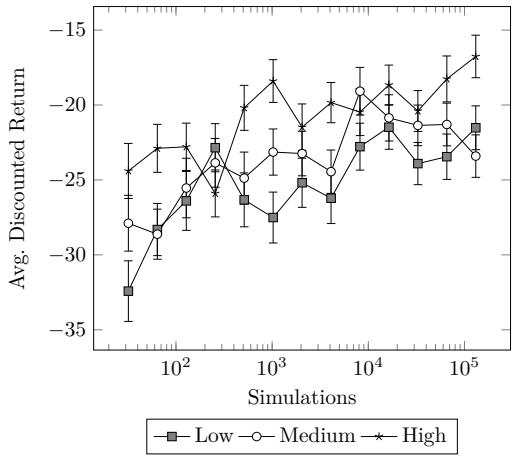
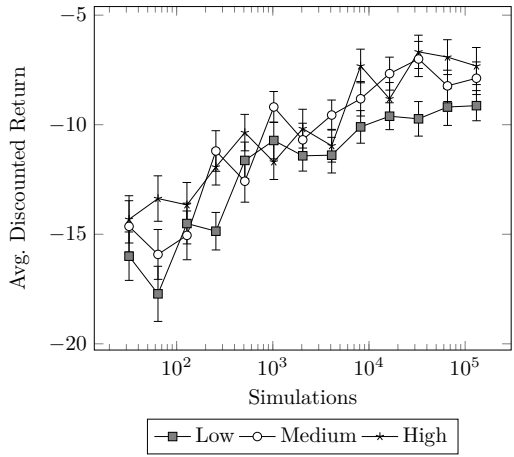
\begin{figure*}[ht]
\centering
	\begin{subfigure}[b]{0.45\textwidth}
	\centering
\pgfplotscreateplotcyclelist{mbw}{%
solid, every mark/.append style={solid, fill=gray}, mark=square*\\%
solid, every mark/.append style={solid, fill=white}, mark=*\\%
solid, every mark/.append style={solid, fill=gray}, mark=star\\%
solid, every mark/.append style={solid, fill=gray}, mark=diamond*\\%
}


\begin{tikzpicture}[scale=0.8]
\begin{semilogxaxis}[
	xlabel={Simulations},
	ylabel={Avg. Discounted Return},
	cycle list name=mbw,
	legend style={at={(0.5,-0.2)},anchor=north},
	legend columns=3
]

\addplot+[error bars/.cd,y dir=both,y explicit,  error bar style={solid}]
coordinates{
(32, -32.42) +- (0, 2.022)
(64, -28.31) +- (0, 1.742)
(128, -26.4) +- (0, 1.964)
(256, -22.85) +- (0, 1.607)
(512, -26.33) +- (0, 1.798)
(1024, -27.51) +- (0, 1.698)
(2048, -25.19) +- (0, 1.639)
(4096, -26.22) +- (0, 1.691)
(8192, -22.78) +- (0, 1.565)
(16384, -21.48) +- (0, 1.474)
(32768, -23.91) +- (0, 1.407)
(65536, -23.45) +- (0, 1.513)
(131072, -21.52) +- (0, 1.462)
};

\addplot+[error bars/.cd,y dir=both,y explicit, error bar style={solid}]
coordinates {
(32, -27.89) +- (0, 1.86)
(64, -28.62) +- (0, 1.67)
(128, -25.54) +- (0, 1.987)
(256, -23.86) +- (0, 1.624)
(512, -24.88) +- (0, 1.738)
(1024, -23.14) +- (0, 1.54)
(2048, -23.24) +- (0, 1.486)
(4096, -24.45) +- (0, 1.445)
(8192, -19.08) +- (0, 1.59)
(16384, -20.86) +- (0, 1.543)
(32768, -21.36) +- (0, 1.356)
(65536, -21.3) +- (0, 1.424)
(131072, -23.41) +- (0, 1.417)
};

\addplot+[error bars/.cd,y dir=both,y explicit, error bar style={solid}]
coordinates {
(32, -24.4) +- (0, 1.836)
(64, -22.89) +- (0, 1.601)
(128, -22.79) +- (0, 1.579)
(256, -25.9) +- (0, 1.566)
(512, -20.19) +- (0, 1.5)
(1024, -18.4) +- (0, 1.426)
(2048, -21.48) +- (0, 1.551)
(4096, -19.84) +- (0, 1.342)
(8192, -20.47) +- (0, 1.571)
(16384, -18.66) +- (0, 1.323)
(32768, -20.4) +- (0, 1.364)
(65536, -18.26) +- (0, 1.532)
(131072, -16.76) +- (0, 1.419)
};

\legend{Low, Medium, High}
\end{semilogxaxis}
\end{tikzpicture}
	\caption{POMCP}
	\label{fig:maintenanceP}
	\end{subfigure}
	\hfill
	\begin{subfigure}[b]{0.45\textwidth}
	\centering
\pgfplotscreateplotcyclelist{mbw}{%
solid, every mark/.append style={solid, fill=gray}, mark=square*\\%
solid, every mark/.append style={solid, fill=white}, mark=*\\%
solid, every mark/.append style={solid, fill=gray}, mark=star\\%
solid, every mark/.append style={solid, fill=gray}, mark=diamond*\\%
}


\begin{tikzpicture}[scale=0.8]
\begin{semilogxaxis}[
	xlabel={Simulations},
	ylabel={Avg. Discounted Return},
	cycle list name=mbw,
	legend style={at={(0.5,-0.2)},anchor=north},
	legend columns=3
]

\addplot+[error bars/.cd,y dir=both,y explicit,  error bar style={solid}]
coordinates{
(32, -16) +- (0, 1.105)
(64, -17.72) +- (0, 1.258)
(128, -14.51) +- (0, 0.9303)
(256, -14.86) +- (0, 0.8564)
(512, -11.63) +- (0, 0.8368)
(1024, -10.72) +- (0, 0.8404)
(2048, -11.42) +- (0, 0.6967)
(4096, -11.39) +- (0, 0.8132)
(8192, -10.1) +- (0, 0.7437)
(16384, -9.612) +- (0, 0.6132)
(32768, -9.732) +- (0, 0.7882)
(65536, -9.193) +- (0, 0.8396)
(131072, -9.131) +- (0, 0.6885)
};

\addplot+[error bars/.cd,y dir=both,y explicit, error bar style={solid}]
coordinates {
(32, -14.64) +- (0, 1.171)
(64, -15.92) +- (0, 1.138)
(128, -15.05) +- (0, 1.113)
(256, -11.2) +- (0, 0.9263)
(512, -12.58) +- (0, 0.9552)
(1024, -9.191) +- (0, 0.7064)
(2048, -10.69) +- (0, 0.7571)
(4096, -9.559) +- (0, 0.685)
(8192, -8.817) +- (0, 0.7871)
(16384, -7.669) +- (0, 0.752)
(32768, -7.001) +- (0, 0.8024)
(65536, -8.227) +- (0, 0.7143)
(131072, -7.879) +- (0, 0.7444)
};

\addplot+[error bars/.cd,y dir=both,y explicit, error bar style={solid}]
coordinates {
(32, -14.32) +- (0, 1.079)
(64, -13.37) +- (0, 1.037)
(128, -13.66) +- (0, 1.023)
(256, -11.93) +- (0, 0.8226)
(512, -10.36) +- (0, 0.8312)
(1024, -11.7) +- (0, 0.8004)
(2048, -10.18) +- (0, 0.8852)
(4096, -10.96) +- (0, 0.7468)
(8192, -7.32) +- (0, 0.7681)
(16384, -8.819) +- (0, 0.7437)
(32768, -6.674) +- (0, 0.76)
(65536, -6.916) +- (0, 0.7956)
(131072, -7.322) +- (0, 0.8459)
};

\legend{Low, Medium, High}
\end{semilogxaxis}
\end{tikzpicture}
	\caption{RAGE}
	\label{fig:maintenanceR}
	\end{subfigure}
\caption{Performance in Maintenance with different worker expertise}
\end{figure*}

The Maintenance scenario was solved successfully by both planners even with very few simulations (100\% of runs reached the terminal state), albeit with poor performance in the case of POMCP. We show the results starting at 32 simulations to enhance the differences between the various levels of worker expertise, with which POMCP appears to struggle (fig. \ref{fig:maintenanceP}).  The worker expertise was set to $0.5$ (low), $0.75$ (medium) and $0.9$ (high); it determines the success probability of container related actions such as ``inspect'' and ``screw'' which makes the worker more or less difficult to predict.  Using RAGE the overall performance including standard error improved significantly and more consistently, with respect to both approximation budget and worker ``expertise'', as seen in fig. \ref{fig:maintenanceR}.

The Assembly domain is much more challenging, with more than $10^{11}$ states for our layout consisting of 6 containers with quantities in $[0,15]$, more than 72 observations (among them the worker actions) and all variations over truck and glue variables. With very little planning budget both planners struggled to reach the terminal state (unsurprisingly), so we start the plot at 256 simulations where we achieved 97\% successful completion.  For 100\% completion, a minimum of 512 and 1024 simulations per step were required by RAGE and POMCP respectively. 

\begin{figure}[ht]
\centering
\pgfplotscreateplotcyclelist{mbw}{%
solid, every mark/.append style={solid, fill=white}, mark=triangle\\%
solid, every mark/.append style={solid, fill=gray}, mark=square*\\%
}


\begin{tikzpicture}[scale=0.8]
\begin{semilogxaxis}[
	xlabel={Simulations},
	ylabel={Avg. Discounted Return},
	cycle list name=mbw,
	legend style={at={(0.5,-0.2)},anchor=north},
	legend columns=2
]
\addplot+[error bars/.cd,y dir=both,y explicit, error bar style={solid}]
coordinates {
(256, -25.62) +- (0, 2.512)
(512, -16.89) +- (0, 1.079)
(1024, -16.82) +- (0, 0.786)
(2048, -13.54) +- (0, 0.7315)
(4096, -11.14) +- (0, 0.5935)
(8192, -7.151) +- (0, 0.2977)
(16384, -7.113) +- (0, 0.302)
(32768, -7.735) +- (0, 0.3543)
(65536, -6.09) +- (0, 0.0475)
(131072, -6.043) +- (0, 0.04617)
};

\addplot+[error bars/.cd,y dir=both,y explicit,  error bar style={solid}]
coordinates{
(256, -29.05) +- (0, 3.304)
(512, -22.08) +- (0, 1.46)
(1024, -16.46) +- (0, 0.6187)
(2048, -13.5) +- (0, 0.4439)
(4096, -12.23) +- (0, 0.3501)
(8192, -12.08) +- (0, 0.4345)
(16384, -11.2) +- (0, 0.3963)
(32768, -11.45) +- (0, 0.4069)
(65536, -11.74) +- (0, 0.4108)
(131072, -11.45) +- (0, 0.4739)
};

\legend{RAGE, POMCP}
\end{semilogxaxis}
\end{tikzpicture}
\caption{Performance in Assembly}
\label{fig:assembly}
\end{figure}
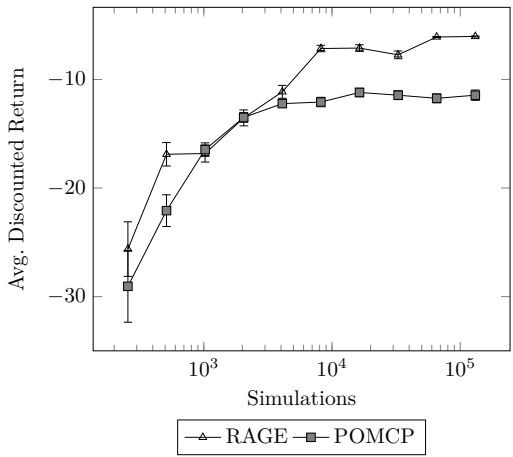

As fig. \ref{fig:assembly} shows, both planners eventually solve the problem but performance increases more rapidly using RAGE, which again obtains a much smaller standard error. We looked at one example RAGE run in detail, and the observer performed particularly well with a combination of information gathering and task-related actions: it restocked empty containers for parts the worker needed, provided the correct glue in advance and never restocked the yellow cabin even though it was empty (a part that is never needed). Some of the apparent performance loss (negative total return) is due to the cost of actions, despite perceiving multiple positive rewards as the worker achieved goal conditions.  An important consideration and potential challenge is attempting to maintain this level of performance across different tasks, or variations of the same task, despite the limitations of planning and acting online such as inaccurate observations and approximation errors.

Finally, table \ref{table:perf} summarizes the best results obtained for each planner in each problem. Although there is still much room for improvement, these results show that active intention recognition tasks can be modeled and solved in an online manner with our proposal and that further, significant improvements may be achieved with more sophisticated frameworks without necessarily relying on highly specific planners or large amounts of domain knowledge.  Based on the perceived peformance, it is possible to improve the lower-bound for worst-case solutions (reflected in the average) as well as the variance that can be achieved with standard, uniformly-random MCTS sampling.

\begin{table}[ht]
\centering
\begin{tabular}{l|c|c}
\hline
\textbf{Problem} & \textbf{Planner} & \textbf{Disc. Return} \\
\hline
\multirow{2}{*}{
Maintenance, Low} & POMCP & -21.52 $\pm$ 1.462 \\
& RAGE & -9.131 $\pm$ 0.689 \\
\hline
\multirow{2}{*}{
Maintenance, Medium} & POMCP & -23.41 $\pm$ 1.417 \\
& RAGE & -7.879 $\pm$ 0.744 \\
\hline
\multirow{2}{*}{
Maintenance, High} & POMCP & -16.76 $\pm$ 1.419 \\
& RAGE & -7.322 $\pm$ 0.846 \\
\hline
\multirow{2}{*}{
Assembly} & POMCP & -11.45 $\pm$ 0.474 \\
& RAGE & -6.043 $\pm$ 0.047 \\
\hline
\end{tabular}
\caption{Performance summary}
\label{table:perf}
\end{table}

\section{Conclusions}
We proposed an approach to active goal and intention recognition, in partially observable domains, that follows a generative approach suitable for online planners that can handle large, unfactored POMDPs. Such contributions to POMDP planning remove the necessity to provide neatly factored problems in advance, and the incorporation of relevance-based planning additionally improves performance without the need for extensive prior knowledge.  Reducing the amount of manual labor and data preparation for specific tasks may be particularly helpful when testing and integrating advances such as these in real-world settings, onboard physical robots.

Our results in AGR may shed some light into a possible direction for improvements, focused on expanding the autonomy of observer agents in active PAIR tasks.  For example, the ability to generate action preferences internally and online, based on a combination of observations and planning, instead of relying on hand-made rules as often happens with special-purpose planners.  We think the structural assumptions made by the RAGE planner are in fact well suited for various robotics tasks, including intention recognition for robotic assistants.  Our work is still in its early stages but preliminary results suggest this is a promising avenue with much potential for improvement.

Future work on active intention recognition will focus on the ability to prioritize observer actions, with respect to hypotheses about target goals. For example if the worker runs out of both cabins and wheels, which one should be restocked first? This topic overlaps with the field of goal reasoning, and is part of a larger effort that includes the integration of various types of planners onboard a physical robot.  Leveraging multiple planners and levels of abstraction, a robotic assistant could reason about goals and intentions modeled as a POMDP and proceed to act in the real-world by decomposing high-level tasks (such as ``bring tool'') into platform-specific actions, for example by introducing an HTN planner.  Additional challenges in automated robotic assistants include developing sufficiently advanced perception, activity and object recognition methods that can produce reliable observations, and the overall difficulty of implementing and testing the resulting platform in a real-world setting.

\section{Acknowledgements}
We'd like to thank our colleagues Oscar Lima, Sebastian Stock and Marc Vinci at PBR in Osnabrück as well as the Smart Factory group in Kaiserslautern. 

This work is part of the InCoRAP project supported by the German Federal Ministry of Education and Research under grant no. 01IW19002. The DFKI Niedersachsen (DFKI NI) is sponsored by the Ministry of Science and Culture of Lower Saxony and the VolkswagenStiftung.

\bibliographystyle{plain}
\bibliography{biblio}

\end{document}